\documentclass[sigconf, screen]{acmart}
\AtBeginDocument{%
  \providecommand\BibTeX{{%
    \normalfont B\kern-0.5em{\scshape i\kern-0.25em b}\kern-0.8em\TeX}}}

\setcopyright{acmcopyright}

\copyrightyear{2023}
\acmYear{2023} 
\setcopyright{acmlicensed}\acmConference[CUI '23]{ACM conference on Conversational User Interfaces}{July 19--21, 2023}{Eindhoven, Netherlands}
\acmBooktitle{ACM conference on Conversational User Interfaces (CUI '23), July 19--21, 2023, Eindhoven, Netherlands}
\acmPrice{15.00}
\acmDOI{10.1145/3571884.3597125}
\acmISBN{979-8-4007-0014-9/23/07}

\usepackage{makecell}
\usepackage{todonotes}
\usepackage{xcolor}
\usepackage{tikz, pgfplots}
\usepackage{graphicx}
\usepackage{booktabs}
\usepackage{amsmath}
\usepackage{multirow}
\usepackage{float}
\usepackage{cancel}

\usepackage{dashundergaps}

\usepackage{pgf}
\usetikzlibrary{arrows,shapes.misc,chains,scopes}
\usepackage{pgfplotstable}
\pgfplotsset{compat=1.15}

\begin{document}

\title[Unified Conversational Models with System-Initiated Transitions between CC and TO Dialogues]{Unified Conversational Models with System-Initiated Transitions between Chit-Chat and Task-Oriented Dialogues}

\author{Ye Liu}
\orcid{0009-0005-4578-769X}
\affiliation{%
  \institution{Mercedes-Benz AG \& Ulm University}
  \city{Sindelfingen \& Ulm}
  \country{Germany}
  }
\email{ye.y.liu@mercedes-benz.com}

\author{Stefan Ultes}
\orcid{0000-0003-2667-3126}
\affiliation{%
  \institution{University of Bamberg}
  \city{Bamberg}
  \country{Germany}}
\email{stefan.ultes@uni-bamberg.de}

\author{Wolfgang Minker}
\orcid{0000-0003-4531-0662}
\affiliation{%
  \institution{Ulm University}
  \city{Ulm}
  \country{Germany}}
\email{wolfgang.minker@uni-ulm.de}

\author{Wolfgang Maier}
\orcid{0000-0001-6396-4956}
\affiliation{%
  \institution{Mercedes-Benz AG}
  \city{Sindelfingen}
  \country{Germany}}
\email{wolfgang.mw.maier@mercedes-benz.com}

\renewcommand{\shortauthors}{Ye Liu, Stefan Ultes, Wolfgang Minker and Wolfgang Maier}


\begin{abstract}

Spoken dialogue systems (SDSs) have been separately developed under two different categories, task-oriented and chit-chat. The former focuses on achieving functional goals and the latter aims at creating engaging social conversations without special goals. Creating a unified conversational model that can engage in both chit-chat and task-oriented dialogue is a promising research topic in recent years. However, the potential ``initiative'' that occurs when there is a change between dialogue modes in one dialogue has rarely been explored. In this work, we investigate two kinds of dialogue scenarios, one starts from chit-chat implicitly involving task-related topics and finally switching to task-oriented requests; the other starts from task-oriented interaction and eventually changes to casual chat after all requested information is provided.
We contribute two efficient prompt models which can proactively generate a transition sentence to trigger system-initiated transitions in a unified dialogue model. One is a discrete prompt model trained with two discrete tokens, the other one is a continuous prompt model using continuous prompt embeddings automatically generated by a classifier. We furthermore show that the continuous prompt model can also be used to guide the proactive transitions between particular domains in a multi-domain task-oriented setting.



\end{abstract}




\maketitle

\section{Introduction}

The discussion about the \textit{initiative} during interaction in spoken dialogue systems (SDSs) dates back to the 1990s. Taking the initiative is expressed as ``taking the conversational lead'' at a more expansive definition \cite{walker1990mixed}. For task-oriented dialogues, the initiative tends to represent ``driving the task'' \cite{smith1994spoken, smith1997effects}. \cite{novick1997mixed} introduces the mixed-initiative interaction and explores that initiative is a multi-factor concept, which includes choice of task, choice of speaker and choice of outcome. Hence, the term "initiative" may be interpreted differently in different dialogue scenarios. In more recent years, \cite{2014fn02} elucidates the challenges of proactiveness in dialogue systems. \cite{sevegnani2021otters} introduces the mixed-initiative topic transitions in a open-domain dialogue system. \cite{liu-etal-2022-system} elaborates on three different types of proactive transitions initiated by a unified dialogue agent. 

\begin{figure*}
\begin{tikzpicture}[scale=1.0]
\footnotesize

\node [] at (0.5, 0.5) {Prepended FusedChat, dialogue mode transition from \textit{chit-chat} to \textrm{task-oriented}};

\node [] at (-3.0, 0) {\includegraphics[width=.03\textwidth]{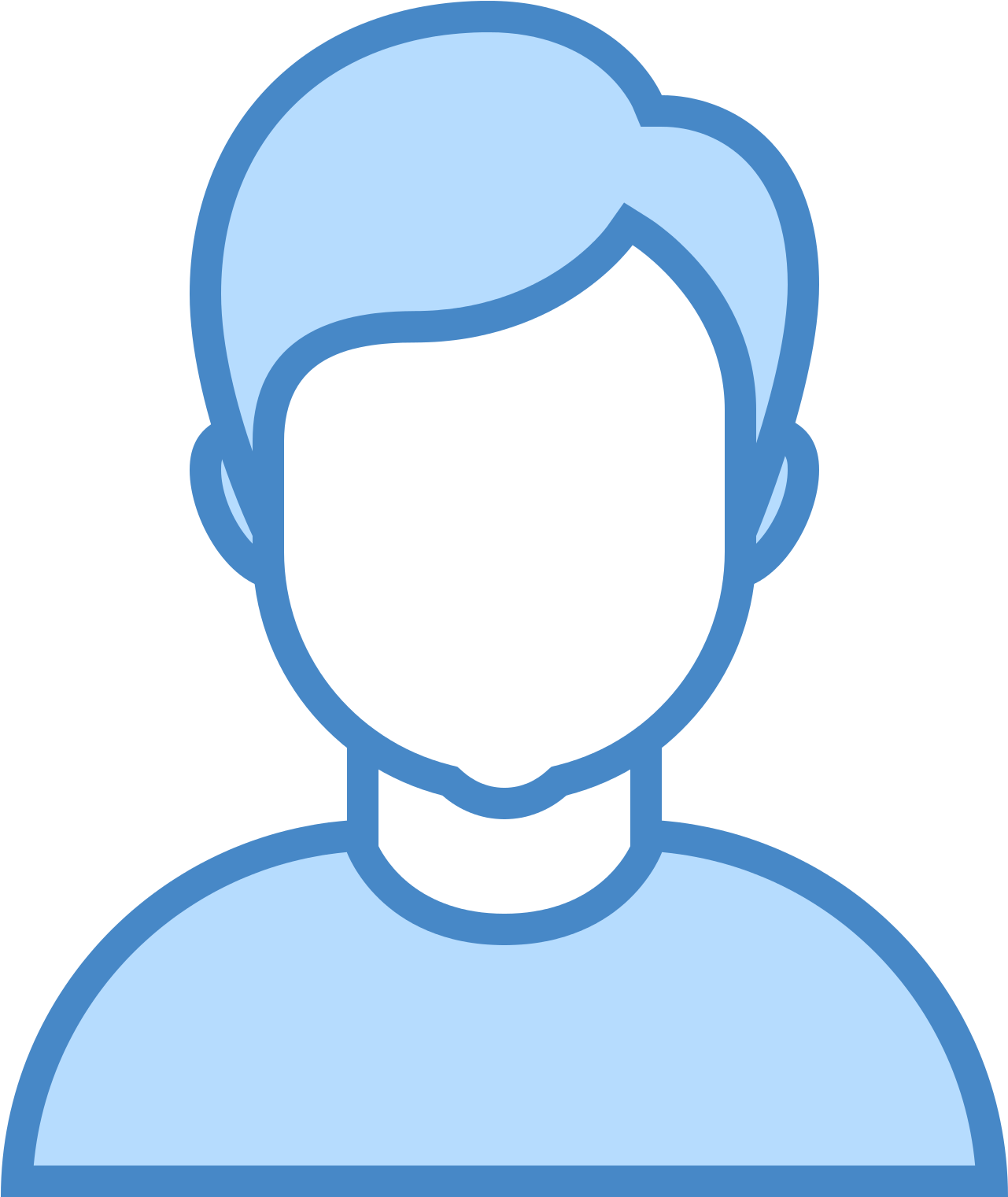}};
\draw [rounded corners] (-2.5, -0.2) rectangle (7.0, 0.2);
\node [] at (2.2, 0.0) {\textit{I will be enrolling in a new school at London Kings Cross next week. I'm so nervous.}};

\node [] at (12, -0.5) {\includegraphics[width=.04\textwidth]{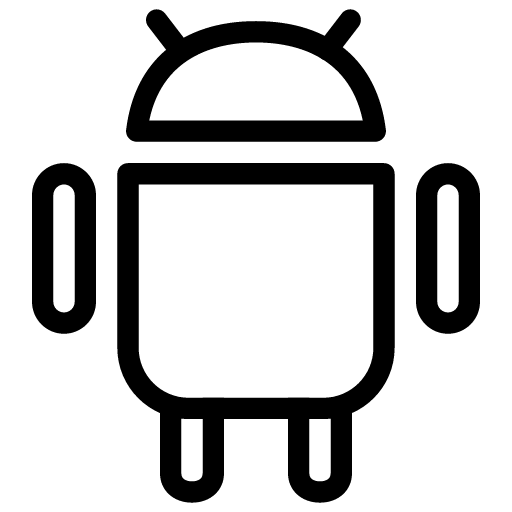}};
\draw [rounded corners] (6.5, -0.7) rectangle (11.5, -0.3);
\node [] at (9.0, -0.5) {\textit{I hope you have fun at your new school.}};

\node [] at (-3.0, -1.0) {\includegraphics[width=.03\textwidth]{pictures/user.png}};
\draw [rounded corners] (-2.5, -1.2) rectangle (7.0, -0.8);
\node [] at (2.2, -1.0) {\textit{Thank you. My family and I will be visiting my school this weekend to see how it's like.}};

\node [] at (12.0, -1.7) {\includegraphics[width=.04\textwidth]{pictures/robot.png}};
\draw [rounded corners] (10.0, -1.7) rectangle (11.5, -1.3);
\node [] at (10.7, -1.5) {\sout{\textit{I see.}}};

\draw [rounded corners] (5.5, -2.2) rectangle (11.5, -1.8);
\node [] at (8.5, -2.0) {\textit{I see.} \textit{\textbf{If you want, I could help you book a train ticket.}}};

\node [] at (-3.0, -2.7) {\includegraphics[width=.03\textwidth]{pictures/user.png}};
\draw [rounded corners] (-2.5, -2.9) rectangle (8.0, -2.5);
\node [] at (2.7, -2.7) {\textrm{I'm trying to find a train that goes from Cambridge to my school. Can you help me book a ticket?}};

\node [] at (12, -3.2) {\includegraphics[width=.04\textwidth]{pictures/robot.png}};
\draw [rounded corners] (4.0, -3.4) rectangle (11.5, -3.0);
\node [] at (7.8, -3.2) {\textrm{I can help with that. Can you tell me what day you will be travelling?}};

\draw[dotted,thick,black] (4.5, -3.5) -- (4.5, -3.7);

\draw[dotted,thick,black] (-3.0, -3.8) -- (12.0, -3.8);
\node [] at (0.5, -4.0) {Appended FusedChat, dialogue mode transition from \textrm{task-oriented} to \textit{chit-chat}};

\node [] at (-3.0, -4.5) {\includegraphics[width=.03\textwidth]{pictures/user.png}};
\draw [rounded corners] (-2.5, -4.7) rectangle (3.5, -4.3);
\node [] at (0.3, -4.5) {\textrm{Do you know where the Parkside Police Station is?}};

\node [] at (12, -5.2) {\includegraphics[width=.04\textwidth]{pictures/robot.png}};
\draw [rounded corners] (5.0, -5.2) rectangle (11.5, -4.8);
\node [] at (8.3, -5.0) {\sout{\textrm{Hello, I can provide the post code for you; it is CB11JG.}}};

\draw [rounded corners] (3.0, -5.7) rectangle (11.5, -5.3);
\node [] at (7.3, -5.5) {\textrm{Hello, I can provide the post code for you; it is CB11JG.} \textrm{\textbf{What happened to you?}}};

\node [] at (-3.0, -6.0) {\includegraphics[width=.03\textwidth]{pictures/user.png}};
\draw [rounded corners] (-2.5, -5.8) rectangle (-0.5, -6.2);
\node [] at (-1.5, -6.0) {\textit{I lost my wallet.}};

\node [] at (12, -6.5) {\includegraphics[width=.04\textwidth]{pictures/robot.png}};
\draw [rounded corners] (7.0, -6.3) rectangle (11.5, -6.7);
\node [] at (9.2, -6.5) {\textit{Don't worry, the police will look into it.}};

\draw[dotted,thick,black] (4.5, -6.6) -- (4.5, -6.8);

\end{tikzpicture}
\caption{Two dialogue examples in FusedChat dataset with our augmented transition sentence at the transition turn. One is a Prepended FusedChat, the other one is a Appended FusedChat. The \textit{italic} part represents the chit-chat interaction, while the \textrm{normal roman} part is task-oriented communication.
Compared with original crossed out response at the transition turn, the initiative dialogue system is able to generate a transition sentence (highlighted in \textbf{bold}) to enable the system-initiated transitions.}
\label{fig: dialogue between user and system shows the arbitrary switch between chit-chat to task-oriented.}
\end{figure*}

In this work, we investigate the initiative in a unified conversational model. Unified SDSs that can reply to both task-oriented and chit-chat requests have been proposed recently \cite{lin2021adapter, young2022fusing, zhao2021unids}, but potential initiatives that emerge as dialogue modes change in one dialogue have not been explored. It is desired that a unified model can not only reply to both chit-chat and task-oriented requests, but also proactively initiate the transition when the user implicitly shows the willingness to switch between chit-chat and task-oriented services. Like dialogue examples shown in Figure \ref{fig: dialogue between user and system shows the arbitrary switch between chit-chat to task-oriented.}, without the bold transition sentences, it is the user who controls the dialogue flow and the dialogue agent is merely providing services mechanically. To enable the proactive capability, the unified SDSs should be developed to recognize the dialogue mode transition and initiate this transition by generating a transition sentence (please see the \textbf{bold} transition sentences in Figure \ref{fig: dialogue between user and system shows the arbitrary switch between chit-chat to task-oriented.}). To be more specific, the following two dialogue scenarios are studied in this paper and the potential benefits are also explained as follows:
\begin{enumerate}
    \item
    The human-machine interaction starts with chit-chat implicitly involving task-related topics and eventually switches to task-related requests, as the first dialogue of Figure \ref{fig: dialogue between user and system shows the arbitrary switch between chit-chat to task-oriented.}.
    The initiative dialogue system consciously anticipates in advance the user need for a specific task-related service, such as ``booking a train ticket'', while kindly asking the user ``If you want, I could help you book a train ticket.''. This is beneficial for commercial dialogue systems to proactively offer or sell their task-related services \citep{chiu2022salesbot, liu-etal-2022-system} at the right time. 
    

    \item
    The human-machine interaction starts with task-oriented requests and eventually switches to casual chat after providing all user needed information, as the second example in Figure \ref{fig: dialogue between user and system shows the arbitrary switch between chit-chat to task-oriented.}.
    Users may have the feeling that they are talking to an acquaintance if the initiative dialogue agent naturally switches to chit-chat interaction after getting all needed task-related information. This can highly improve user interaction experience \cite{liu-etal-2022-system}, even provide mental support \cite{ring2013addressing}.
\end{enumerate}


In summary, the main contributions of this paper are as follows:
    
    

\begin{itemize}
    \item 
    We utilize the original FusedChat dataset, where each dialogue includes two dialogue modes with chit-chat and task-oriented parts being interdependent (please see dialogue examples in Figure \ref{fig: dialogue between user and system shows the arbitrary switch between chit-chat to task-oriented.}) and adopt the pre-trained GPT-2 \citep{radford2019language}
    to build a unified dialogue model
    (section \ref{sec: unified generation model}),
    which can generate both chit-chat and task-oriented responses.
    This is the basis for the subsequent work.
    
    \item
    We artifically augment transition sentences (section \ref{sec: dataset collection}) for $592$
    FusedChat \citep{young2022fusing} dialogues, which include Appended FusedChat and Prepended FusedChat.
    These augmented transition sentences are further applied for the discrete and continuous prompt learning.

    \item 
    We leverage discrete and continuous prompt learning
    (section \ref{sec: Activate the Initiatitve of Unified Model})
    to efficiently extend the proactiveness capability to the unified model
    for generating a transition sentence at the transition turn.
\end{itemize}

The remainder of this paper is structured as follows: Section~\ref{sec: related work} shows the related works of our research. Section~\ref{sec: dataset collection} introduces the human annotation task on transition sentences collection. Section \ref{sec: unified generation model} briefly describes the unified model that can reply to both chit-chat and task-oriented requests. Section \ref{sec: Activate the Initiatitve of Unified Model} mainly introduces the discrete and continuous prompt learning to activate the initiative transitions of the unified model. Section \ref{sec: Results Comparison and Use Case Study} elaborates on the performance of our proposed models via automatic metric scores and use case study. Section~\ref{sec:Conclusion and Future Work} concludes this work and outlines future research.

\section{Related Work}
\label{sec: related work}

There are several works on using chit-chat utterances to augment task-oriented dialogue datasets. For instance, \cite{zhao2017generative} artificially augmented a task-oriented dataset with randomly sampled open-domain utterances. It demonstrated that the model trained on the augmented data can improve the out-of-domain recovery performance for the task-oriented system. \cite{xu2021augnlg} introduced a novel data augmentation approach to automatically create MR-to-Text data from open-domain texts. \cite{sun2021adding} proposed a method for adding contextually relevant chit-chat to enhance the engaging and social of task-oriented dialogues. However, these models only concern the performance of the underlying task-oriented dialogue model.

\begin{table*}[h!!]
\centering
\caption{Statistics of human augmented dialogues. Because $470$ dialogues are multi-domain, the sum of domain distribution is more than $592$.} 
\begin{tabular}{ccc}
\toprule
train & test & valid\\
478 & 59 & 55 \\
\bottomrule
\end{tabular}
\quad
\begin{tabular}{cc}
\toprule
Prepended & Appended \\
248 & 344 \\
\bottomrule
\end{tabular}
\quad
\begin{tabular}{ccccccc}
\toprule
restaurant & hotel & attraction & train & taxi & police & hospital \\
286  & 270 & 244 & 234  & 164 & 16 & 11\\
\bottomrule
\end{tabular}

\label{tab: statistic of human annotated dialogues}
\end{table*}

Several previous works combined chit-chat and task-oriented datasets to train a unified dialogue model. For instance, \cite{shuster2020dialogue} introduced the dodecaDialoguen task, to assemble important aspects of an engaging conversational agent into a single collection by leveraging $12$ tasks. The Adapter-Bot presented in \cite{lin2021adapter} utilized multiple adapter layers with the pre-trained DialoGPT model to activate new response skills and styles. \cite{zhao2021unids} proposed a dialogue model for training chit-chat and task-oriented in a unified data schema, which both include belief states, representation of dataset results, and system acts. However, these models simply fuse chit-chat dialogue and task-oriented dialogue into one model and do not consider the dependency between different types of dialogues in the multi-turn setting. In contrast, all dialogues in FusedChat released in \cite{young2022fusing} include both chit-chat and task-oriented turns, and treats them as parallel dialogue modes of equal importance. While in this way, the contextual dependency between these two dialogue modes is taken into account, the potential system initiative that occurs during switchover has not been discussed.
Our work, on the other hand, leverages FusedChat to explore the proactive transitions between these two dialogue modes guided by dialogue systems rather than users. \cite{chiu2022salesbot} recently proposed SalesBot and investigated the conversations starting from open-domain social chatting and then gradually transitioning to task-oriented purposes. However, our work not only explores the transition from chit-chat to task-oriented, but also from task-oriented to chit-chat. \cite{liu-etal-2022-system} recently also discussed three kinds of initiative transitions in a unified dialogue system.

The recent achievement in prompt learning \cite{liu2021pre, li2022personalized} enables researchers to efficiently adapt a given task to pre-trained models rather than modifying the structure of models \cite{lester2021power}, which also inspired our work. The prompt learning is generally categorized into discrete prompt learning, where the prompt is natural language string and was used in GPT-3 \cite{brown2020language}; and continuous/soft prompt learning, where the prompt is encoded in a continuous embedding space. Our work utilizes both discrete and continuous prompt learning to efficiently activate the system-initiated transitions for the unified conversational model.

\section{Transition Sentences Collection}
\label{sec: dataset collection}

In this work, we utilize the FusedChat dataset \cite{young2022fusing}, where human augmented open-domain dialogues are prepended and appended to the dialogues of the popular task-oriented dataset MultiWOZ \cite{budzianowski2018multiwoz, DBLP:journals/corr/abs-2104-00773}. In Appended FusedChat, the appended open-domain sentences are content-dependent on the preceding task-oriented interaction. In Prepended FusedChat, the prepended open-domain dialogue must retain a slot value mentioned in the first turn of the succeeding task-oriented dialogue to enable the inter-mode dependency. Hence, every dialogue in FusedChat includes two dialogue modes with chit-chat and task-oriented parts being interdependent. As the dialogue examples in Figure \ref{fig: dialogue between user and system shows the arbitrary switch between chit-chat to task-oriented.}, most FusedChat dialogues do not take the dialogue mode transitions initiated by system into account, so we manually augment transition sentences at transition turns (highlighted in \textbf{bold} in Figure \ref{fig: dialogue between user and system shows the arbitrary switch between chit-chat to task-oriented.}) for $592$ FusedChat dialogues and then utilize these augmented dialogues to train the initiative models (section \ref{sec: Activate the Initiatitve of Unified Model}). In the human augmentation task, we employ two Master students with computational linguistics background as annotators. To validate and improve the annotation schemes, we have defined the following annotation guidelines:
\begin{enumerate}
  \item 
  When creating a transition sentence, both chit-chat and task-oriented contextual segments have to be considered.
  Therefore, augmented transition sentences must be semantically and contextually reasonable.

  \item
  In Appended FusedChat dialogues (from task-oriented to chit-chat), sentences with patterns like ``Do you need anything else?'', ``What else can I do for you?'' and ``Is there anything else I can do for you?'' are commonly occurring in the last system turn of task-oriented interaction. To better reflect system-initiated transitions to the chit-chat interaction, these generic responses have to be removed firstly and an informative transition sentence is rewritten.
  The chit-chat in Appended FusedChat dialogues usually starts with a declarative sentence containing user’s personal thoughts. In this case, using an interrogative as transition sentence can better induce to the chit-chat interaction (see the second dialogue in Figure \ref{fig: dialogue between user and system shows the arbitrary switch between chit-chat to task-oriented.}). 
  
  \item
  In Prepended FusedChat dialogues (from chit-chat to task-oriented), since the prepended chit-chat already includes one slot that exists in the first user utterance of the succeeding task-oriented interaction, the annotated transition sentences must be more task-oriented rather than open-domain sentences for smooth switching to task-oriented interaction (see the transition sentence of the fist dialogue in Figure \ref{fig: dialogue between user and system shows the arbitrary switch between chit-chat to task-oriented.}). At the same time, the annotators are instructed to avoid generic transition sentences which offer advice or provide further help without specific intention, such as ``do you need some recommendations?''.

\end{enumerate}

To increase reliability and establish agreement among annotators, two annotators independently augmented $50$ common samples in the first step and conclude annotation guidelines to follow in the follow-up annotation section. We finally annotate a total of $592$ dialogues, which include both Prepended and Appended FusedChat examples, cover different task domains and span over train/test/valid dataset. Some Appended dialogues need to be relabelled due to ambiguous dialogue mode labels. Hence, we re-label these dialogue modes and write a transition sentence at the transition turn for those dialogues. That is why the number of annotated Appended FusedChat is larger than Prepended in Table \ref{tab: statistic of human annotated dialogues}, which shows statistics of the human augmented dialogues.


\begin{figure*}
\centering
\begin{tikzpicture}[scale=0.9]

\node at (-2.5, 4)[align=center]  {\underline{Unified Generation Model: }};

\draw[->,thick] (8.0, 2.5) -- (8.0, 3);
\node at (7.0, 3.1)[align=center]  {\textit{i hope you have fun at your new school. [END]}};
\node at (6.0, 3.6)[align=center]  {{hello, i can provide the post code for you; it is CB11JG. [END]}};

\draw[thick,rounded corners]   (5, 1.9) rectangle (10, 2.5);
\node at (7.5, 2.2)[align=center]  {{pre-trained GPT-2}};

\draw[->,thick] (8.0, 1.4) -- (8.0, 1.9);
\node at (4.0, 1.3)[align=center]  {\textit{[USER] i will be enrolling in a new school at london kings cross ... [SYSTEM]}};
\node at (4.0, 0.8)[align=center]  {{[USER] do you know where the parkside ...? police\{inform (post=CB11JG)\} [SYSTEM]}};

\draw[dotted,thick,black] (-5.0, 0.6) -- (11, 0.6);

\node at (-2.5, 0.3)[align=center]  {\underline{Discrete Prompt Model: }};

\draw[->,thick] (8.0, -1.3) -- (8.0, -0.8);
\node at (5.0, -0.2)[align=center]  {{hello, i can provide ... it is CB11JG.} \textbf{[TRANSITION] what happened to you? [END]}};
\node at (7.2, -0.7)[align=center]  {\textit{i hope you have fun at your new school. [END]}};

\draw[thick,rounded corners]   (5, -1.9) rectangle (10, -1.3);
\node at (7.5, -1.6)[align=center]  {{unified generation model}};

\draw[->,thick] (8.0, -2.4) -- (8.0, -1.9);
\node at (3.0, -2.5)[align=center]  {\textit{[CHIT-CHAT] [NORMAL-TURN] [USER] i will be enrolling in a new school ... [SYSTEM]}};
\node at (3.0, -3.0)[align=center]  {{[TASK-ORIENTED] [TRANSITION-TURN] [USER] do you know where the parkside ... [SYSTEM]}};

\draw[dotted,thick,black] (-5.0, -3.2) -- (11, -3.2);

\node at (-2.5, -3.5)[align=center]  {\underline{Continuous Prompt Model: }};

\draw[->,thick] (8.0, -5.1) -- (8.0, -4.6);
\node at (5.3, -4.5)[align=center]  {\textit{i hope you have fun at your new school. [END]}};
\node at (4.8, -4.0)[align=center]  {{hello, i can provide ...} \textbf{[TRANSITION] what happened to you? [END]}};

\draw[thick,rounded corners]   (5, -5.7) rectangle (10, -5.1);
\node at (7.5, -5.4)[align=center]  {{discrete prompt model}};

\draw[->,thick] (8.0, -6.2) -- (8.0, -5.7);
\node at (6.0, -6.3)[align=center] {\textit{[USER] i will be enrolling .. [SYSTEM]}};
\node at (6.0, -6.8)[align=center] {{[USER] do you know where ... [SYSTEM]}};

\draw[->, thick] (1.6, -5.4) -- (5, -5.4);

\draw[thick,rounded corners]   (-2.0, -5.7) rectangle (1.6, -5.1);
\node at (-0.2, -5.4)[align=center]  {{LSTM bridge layer}};

\draw[->,rounded corners] (-3.5, -6.4) -- (-3.5, -5.9) -- (-2.8, -5.9)-- (-2.8, -5.4) -- (-2.0, -5.4);
\draw[->,rounded corners] (-2.0, -6.4) -- (-2.0, -5.9) -- (-2.8, -5.9)-- (-2.8, -5.4) -- (-2.0, -5.4);

\draw[thick,rounded corners]   (-4.1, -7.0) rectangle (-2.9, -6.4);
\node at (-3.5, -6.7)[align=center]  {{CCTO}};
\draw[thick,rounded corners]   (-2.6, -7.0) rectangle (-1.4, -6.4);
\node at (-2.0, -6.7)[align=center]  {{TTNT}};

\draw[->,rounded corners] (-2.8, -8.0) -- (-2.8, -7.5) -- (-3.5,-7.5) -- (-3.5, -7.0);
\draw[->,rounded corners] (-2.8, -8.0) -- (-2.8, -7.5) -- (-2.0,-7.5) -- (-2.0, -7.0);

\draw[thick,rounded corners]   (-3, -8.6) rectangle (2, -8.0);
\node at (-0.5, -8.3)[align=center]  {{pre-trained RoBERTa}};

\draw[->,thick] (-0.5, -9.1) -- (-0.5, -8.6);
\node at (-1.0, -9.2)[align=center] {\textit{[CLS] i will be enrolling...  [SEP]}};
\node at (-1.0, -9.7)[align=center] {{[CLS] do you know where ... [SEP]}};

\end{tikzpicture}
\caption{Architecture of the unified generation model adapted from the pre-trained GPT-2, the discrete prompt model adapted from the unified generation model, the continuous prompt model adapted from the discrete prompt model with fine-tuned RoBERTa classifiers. Every model is attached both task-oriented and chit-chat (highlighted in \textit{italic}) data example. The transition sentences are highlighted in \textbf{bold}.
Compared with the unified model, two additional discrete prompt tokens (in Table \ref{tab: Different combination of two prompt tokens to activate the different mode generation.}) are prepended to the input of the discrete prompt model.
In the continuous prompt model, the two continuous prompt tokens are converted from the RoBERTa classifiers outputs through LSTM bridge layer.
}
\label{fig: architecture of unified model, plus discrete prompt learning and continuous prompt learning with pre-trained classifiers.}
\end{figure*}
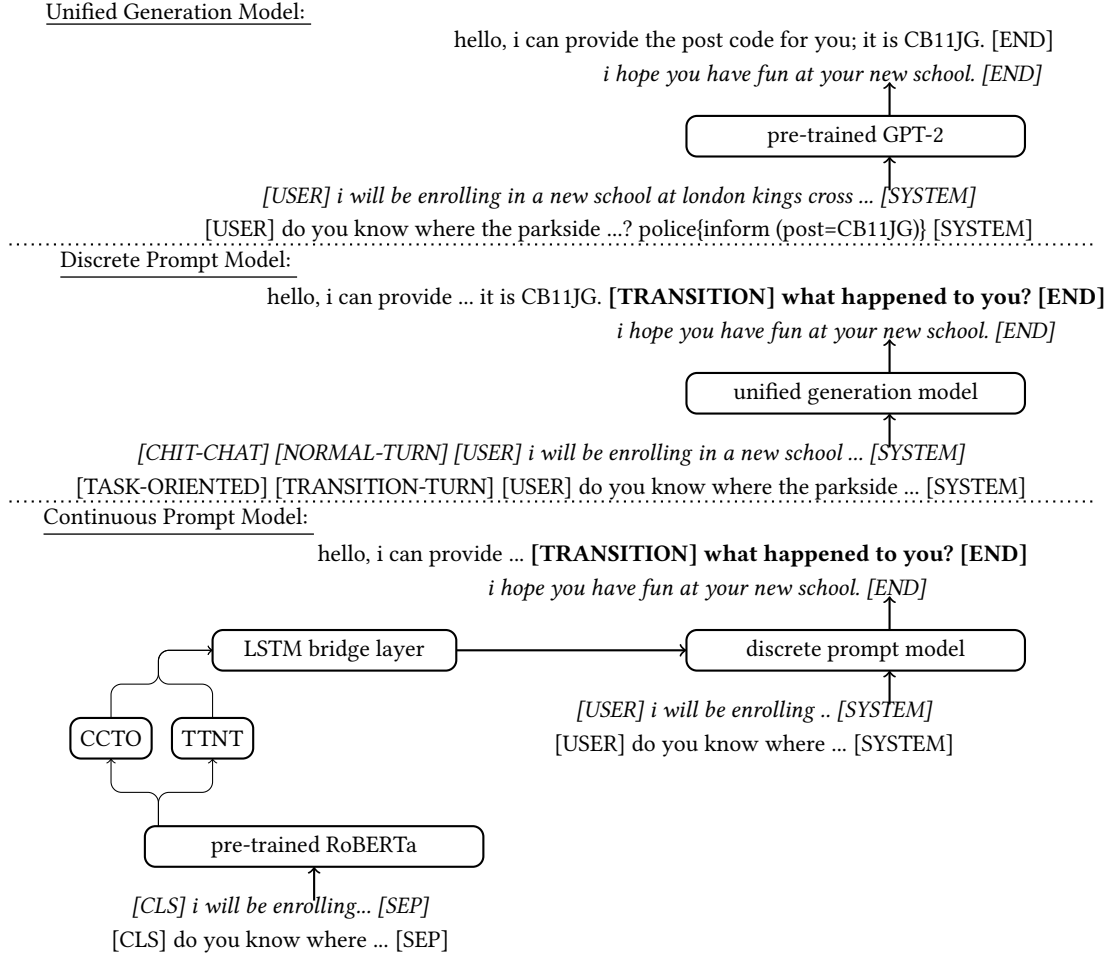

\section{Unified Generation Model}
\label{sec: unified generation model}

We build a unified generation model in the first step and the initiative systems are adapted from this unified model. We tackle the unified generation problem through fine-tuning conditional GPT-2 \cite{radford2019language}. Given the FusedChat dataset $\mathcal{D} = \{(\emph{u}_{n}, \emph{d}_{n}, \emph{r}_{n})_{n=1}^{N}, (\emph{u}_{m}, \emph{r}_{m})_{m=1}^{M} \} $ with $N$ task-oriented samples and $M$ chit-chat samples, the goal is to build a unified generation model parameterized by $\theta$ to be able to respond to both chit-chat and task-oriented requests, as shown in the Equation \ref{equ: GPT-2 for CC and TO generation},
\begin{equation}
\label{equ: GPT-2 for CC and TO generation}
    p_{\theta}(\emph{r}) = 
    \begin{cases}
      \prod_{t=1}^{T} p_{\theta}(r_{t}|r_{<t}, \emph{u}, \emph{d}) & \text{if task-oriented} \\
      \prod_{t=1}^{T} p_{\theta}(r_{t}|r_{<t}, \emph{u}) & \text{if chit-chat}
    \end{cases}
\end{equation}
where $r_{<t}$ indicates all tokens before $t$. The $\emph{u}$ represents the dialogue context; $\emph{d}$ means the dialogue actions only exist in the task-oriented data and $\emph{r}$ is the system response which includes $(r_{1}, ... r_{t}, ...)$ tokens with length $T$. The $\theta$ is optimized via maximizing the loglikelihood (MLE) of the conditional probabilities
in Equation \ref{equ: GPT-2 for CC and TO generation} over the entire task-oriented and chit-chat dataset:
\begin{equation}
\label{equ: objective function of unified model}
\begin{aligned}
    \mathcal{L}_{\theta}(\mathcal{D}) = \sum_{n=1}^{N} \sum_{t=1}^{T_{n}} \log{p_{\theta}(r_{t,n}|r_{<t,n}, \emph{u}_{n}, \emph{d}_{n})} \\ 
    + \sum_{m=1}^{M} \sum_{t=1}^{T_{m}} \log{p_{\theta}(r_{t,m}|r_{<t,m}, \emph{u}_{m})}
\end{aligned}
\end{equation}

We utilize the FusedChat for the training of unified generation model. The train/validation/test split of FusedChat dataset is aligned with MultiWOZ dataset, hence we also use the same split in this paper.
During the GPT-2 fine-tuning (please see the unified generation model in Figure \ref{fig: architecture of unified model, plus discrete prompt learning and continuous prompt learning with pre-trained classifiers.}), we add $\text{[USER]}$ and $\text{[SYSTEM]}$ to the GPT-2 tokenizer to distinguish user utterances from system responses. Only one preceding user utterance is used as the dialogue context for response generation. The behind reasons are, firstly, the context size analysis in \cite{liu-etal-2021-context} demonstrated that the immediately preceding user utterance contains most useful information for contextual response generation; secondly, using a smaller context is more memory-efficient for training. During training, the learning rate is $5\mathrm{e}{-5}$, batch size is $16$. The best model is saved at epoch $5$ with early stopping. We mix top-K sampling and top-p (nucleus) sampling \cite{holtzman2019curious} for decoding. We apply top-K of $5$ and top-p of $0.9$ for chit-chat response generation and top-K of $10$ and top-p of $0.5$ for task-oriented response generation respectively\footnote{In our experiment, we found that decoding with lower top-p can generate task-oriented responses that have better automatic metric scores.}. The discrete and continuous prompt model (section \ref{sec: Activate the Initiatitve of Unified Model}) follow with the same decoding strategy as the unified model.

\begin{table*}[h!!]
\centering
\caption{Four combinations of two special prompt tokens to activate different generation modes in the discrete prompt model. The CCTO is an acronym for Chit-Chat and Task-Oriented; while TTNT stands for Transition-Turn and Normal-Turn.}
\scalebox{0.9}{
\begin{tabular}{ccc}
\toprule
    first CCTO prompt & second TTNT prompt & response generation \\
\midrule
    $\text{[CHIT-CHAT]}$ & $\text{[TRANSITION-TURN]}$ & a chit-chat response plus a transition sentence for the proactive transition to task-oriented interaction \\
    $\text{[CHIT-CHAT]}$ & $\text{[NORMAL-TURN]}$ & a normal chit-chat response without a transition sentence\\
    $\text{[TASK-ORIENTED]}$ & $\text{[TRANSITION-TURN]}$ & a task-oriented response plus a transition sentence for the proactive transition to chit-chat interaction \\
    $\text{[TASK-ORIENTED]}$ & $\text{[NORMAL-TURN]}$ & a normal task-oriented response without a transition sentence \\
\bottomrule
\end{tabular}}
\label{tab: Different combination of two prompt tokens to activate the different mode generation.}
\end{table*}

\section{System-initiated Unified Model}
\label{sec: Activate the Initiatitve of Unified Model}

We further extend the initiative capability of the unified generation model that is able to distinguish between chit-chat and task-oriented responses, while proactively initiating transitions between these two dialogue modes through generating transition sentences. Given the efficient performance of prompt learning \cite{liu2021pre, li2022personalized, brown2020language}, we utilize only $\textit{TWO}$ tokens as the prompt to enable the proactive feature. The first prompt token indicates that the dialogue system should generate a chit-chat or task-oriented response, the second prompt token indicates whether the system should generate a transition sentence to proactively initiate the dialogue mode transition.

We propose two system-initiated generation models. One is utilizing two discrete tokens as the prompt, the other one is using two continuous embeddings as the prompt. The former discrete prompt model and the latter continuous prompt model are introduced in detail in this section.


\subsection{Discrete Prompt Model}
\label{subsec: Discrete Prompting}

Adapted from the unified generation model, we first add the following special tokens to the GPT-2 tokenizer for discrete prompt learning.
\begin{itemize}
  \item [1)]
  The $\text{[CHIT-CHAT]}$ token implies the system should generate a chit-chat response.

  \item [2)]
  The $\text{[TASK-ORIENTED]}$ implies the system should generate a task-oriented response.
  
  \item [3)]
  The $\text{[TRANSITION-TURN]}$ token implies the dialogue system to generate a transition sentence for proactively initiating the dialogue mode transition.

  \item [4)]
  The $\text{[NORMAL-TURN]}$ token implies the system to stay in the original dialogue mode and generate the same type of dialogue as preceding context without transition sentences.
  
  \item [5)]
  The $\text{[TRANSITION]}$\footnote{In addition to, $\text{[TRANSITION]}$ is also used to evaluate the performance of initiative models for generating transition sentences in section \ref{sec: Results Comparison and Use Case Study}.} token is a special token inserted into the response at the transition turn to separate normal response and transition sentence.
\end{itemize}
During discrete prompt learning, we prepend two special tokens\footnote{The number of discrete prompt tokens, TWO, is aligned with the continuous prompt model, where TWO outputs of RoBERTa classifiers are converted to the continuous prompt through LSTM bridge layer (section \ref{subsec: Continuous Prompting}).} as discrete prompt to the input to indicate what kind of response the dialogue system should generate, a chit-chat or task-oriented response, with or without a transition sentence to initiate the dialogue mode transitions (see the discrete prompt model in Figure \ref{fig: architecture of unified model, plus discrete prompt learning and continuous prompt learning with pre-trained classifiers.}). The four prompt combinations representing different types of response generation are shown in Table \ref{tab: Different combination of two prompt tokens to activate the different mode generation.}. The CCTO is an acronym for Chit-Chat and Task-Oriented; while TTNT stands for Transition-Turn and Normal-Turn in this paper.

As the prompt tokens in Table \ref{tab: Different combination of two prompt tokens to activate the different mode generation.} are newly added in this step, we need to activate them through continually training the unified model with human augmented dialogues (section \ref{sec: dataset collection}). For the discrete prompt learning, only dialogue generations at transition turns as training dataset, which includes a normal response without a transition sentence activated by $\text{[NORMAL-TURN]}$ at the second prompt, and a normal response with transition sentence activated by $\text{[TRANSITION-TURN]}$. The Figure \ref{fig: architecture of unified model, plus discrete prompt learning and continuous prompt learning with pre-trained classifiers.} visualizes input examples during discrete prompt training. For the system-initiated unified model, we try to take more dialogue context into account for generating a contextual transition sentence and finally maximal $3$ dialogue turns are used for memory-efficient training. Adapted from the unified GPT-2 model (section \ref{sec: unified generation model}), batch size for training of the discrete prompt model is $16$, learning rate is $5\mathrm{e}{-5}$, and the best discrete prompt model is saved at epoch $4$ with early stopping.


\subsection{Continuous Prompt Model}
\label{subsec: Continuous Prompting}

In discrete prompt learning, we have to artificially add two prompt tokens to each dialogue input to trigger different generation modes. To realize a fully automated process, we propose a continuous prompt model where the RoBERTa \cite{liu2019roberta} classifiers predict generation modes and the outputs are then automatically converted to two continuous prompts by a long-short term memory network (LSTM) \cite{hochreiter1997long} bridge layer (see the continuous prompt model in Figure \ref{fig: architecture of unified model, plus discrete prompt learning and continuous prompt learning with pre-trained classifiers.}). Therefore, the continuous prompt model learns proactive transitions directly from dialogue context and the generated latent high-dimensional continuous prompts can also help to discover other transition possibilities besides dialogue mode transitions.


\subsubsection{RoBERTa Classifiers}
\label{subsubsec: RoBERTa Classifier}

To achieve full automation and to learn initiative transitions directly from dialogue context, we utilize the pre-trained RoBERTa\footnote{We also tried the pre-trained BERT for this classification task. The experiment results show similar performance between BERT and RoBERTa. However, the works in \cite{rothe2020leveraging} and \cite{10.1007/978-981-19-5538-9_5} demonstrate that the combination of pre-trained RoBERTa and GPT-2 works better. Hence, we finally opted for the pre-trained RoBERTa.} to predict generation modes given dialogue history. The outputs of the trained RoBERTa classifiers are converted into two continuous prompts to replace the word embeddings of the two discrete prompt tokens in the discrete prompt model.

The same FusedChat dataset as section \ref{sec: unified generation model} is used for fine-tuning RoBERTa classifiers. We define the training dataset $\mathcal{D}^{'} = \{(\emph{h}_{l}, \emph{y}_{l}^{ccto}, \emph{y}_{l}^{ttnt})_{l=1}^{L} \}$ with $L$ data samples here, where \emph{h} represents dialogue history, $\emph{y}^{ccto}$ and $\emph{y}^{ttnt}$ are the ground-truth label of CCTO and TTNT classifiers respectively. Given dialogue context \emph{h}, the pre-trained RoBERTa \cite{liu2019roberta} returns a sequence of contextualized vectors:
\begin{equation}
\label{equ: roberts cls output}
    \emph{v}_{\text{[CLS]}}, \emph{v}_{h_{1}}, ... = \emph{RoBERTa}(\emph{h})
\end{equation}
Then two classifier layers, CCTO classifier in Equation \ref{equ: CCTO classifiers} and TTNT classifier in Equation \ref{equ: TTNT classifiers}, are separately added on the pooled output of [CLS] token, $\emph{v}_{\text{[CLS]}}$.
\begin{equation}
\label{equ: CCTO classifiers}
\begin{aligned}
    \emph{p}^{ccto} & = \mathbf{W_{0}}^{ccto} \emph{v}_{\text{[CLS]}} + \mathbf{b_{0}}^{ccto} 
    \\
    \hat{\emph{y}}^{ccto} & = \mathbf{W_{1}}^{ccto} \textit{Dropout}(\emph{p}^{ccto}) + \mathbf{b_{1}}^{ccto}s
\end{aligned}
\end{equation}


\begin{equation}
\label{equ: TTNT classifiers}
\begin{aligned}
    \emph{p}^{ttnt} & = \mathbf{W_{0}}^{ttnt} \emph{v}_{\text{[CLS]}} + \mathbf{b_{0}}^{ttnt}  
    \\
    \hat{\emph{y}}^{ttnt} & = \mathbf{W_{1}}^{ttnt} \textit{Dropout}(\emph{p}^{ttnt}) + \mathbf{b_{1}}^{ttnt} 
\end{aligned}
\end{equation}

The binary CCTO classifier is responsible for predicting the dialogue mode, i.e., that the dialogue system should generate a chit-chat or task-oriented response (same meaning as the fist prompt token in discrete prompt model, see Table \ref{tab: Different combination of two prompt tokens to activate the different mode generation.}); the binary TTNT classifier is responsible for predicting whether the dialogue model should generate a transition sentence to trigger the system-initiated transition, i.e., whether the generation is at a transition turn or normal turn (same meaning as the second prompt token in discrete prompt model).

In Equation \ref{equ: CCTO classifiers} and \ref{equ: TTNT classifiers}, the $\emph{p}^{ccto}$ and $\emph{p}^{ttnt}$ are the pooled output of classifiers and will be fed into LSTM bridge layer (section \ref{subsubsec: LSTM Bridge Layer}) for generating two continuous prompts\footnote{We choose the pooled outputs rather than binary classifier outputs to feed into LSTM bridge layer because the pooled outputs have the same embedding size as GPT-2 embedding layer, $768$.}. The dropout layer in classifier layers has the same dropout rate as RoBERTa model $0.1$. The $\hat{\emph{y}}^{ccto}$ and $\hat{\emph{y}}^{ttnt}$ are the output of binary CCTO and TTNT classifier respectively. The $\mathbf{W}$ and $\mathbf{b}$ are trainable parameters and updated during the training of RoBERTa classifiers. All parameters of RoBERTa classifiers $\theta^{'}$ are jointly trained via minimizing cross-entropy losses of both CCTO and TTNT, as the Equation \ref{equ: objective function of RoBERTa} shown:
\begin{equation}
\label{equ: objective function of RoBERTa} 
\begin{aligned}
    \mathcal{L}_{\theta^{'}}(\mathcal{D}^{'}) = - \sum_{l=1}^{L} \sum_{c=1}^{2} \emph{y}_{lc}^{ccto} \log{\hat{\emph{y}}_{lc}^{ccto}} \\ 
    - \sum_{l=1}^{L} \sum_{c=1}^{2} \emph{y}_{lc}^{ttnt} \log{\hat{\emph{y}}_{lc}^{ttnt}}  
\end{aligned}
\end{equation}
\emph{c} is in $[1, 2]$, because CCTO and TTNT classifier are binary classifier.

The RoBERTa is trained independently from the unified GPT-2 generation model. The input for RoBERTa is the complete dialogue history with maximal $256$ tokens limitation. The [CLS] token is inserted to the first position and [SEP] token is used to separate user utterances and system responses in the input. The RoBERTa is fine-tuned using AdamW optimizer \citep{loshchilov2017decoupled} with a learning rate of $5\mathrm{e}{-5}$. The batch size is $60$ and the best model is saved at epoch $4$ with early stopping. Table \ref{tab: the performance of RobERTa classifier.} shows the performance of fine-tuned RoBERTa for the binary CCTO and TTNT classifier.

\begin{table}
\centering
\caption{Performance of fine-tuned RoBERTa classifiers, the binary CCTO classifier and the binary TTNT classifier. Only one turn in multi-turn FusedChat dialogues is labelled as the transition turn, i.e., the training dataset for the TTNT classifier is highly imbalanced. Hence, we compute the weighted metrics for TTNT classifier.}
\scalebox{1.0}
{\begin{tabular}{ccccc}
\toprule
     & accuracy & recall & precision & f1 \\
\midrule
    CCTO & 98.89 & 98.48 & 98.83 & 98.65 \\
    TTNT (weighted) & 92.05 & 92.05 & 91.28 & 91.52 \\

\bottomrule
\end{tabular}}
\label{tab: the performance of RobERTa classifier.}
\end{table}

\subsubsection{LSTM Bridge Layers}
\label{subsubsec: LSTM Bridge Layer}
After RoBERTa fine-tuning, we further project the high-dimensional outputs of RoBERTa classifiers to the GPT-2 decoder space. The two pooled outputs of CCTO and TTNT classifiers are first converted to two continuous prompt embeddings by the LSTM bridge layers and then used to replace the word embeddings of the two discrete prompt tokens of the discrete prompt model to activate different generation modes. In addition, the latent high-dimensional continuous prompts can learn other potential transition possibilities beyond the four types of generation modes (see Table \ref{tab: Different combination of two prompt tokens to activate the different mode generation.}) in the discrete prompt model.

Inspired from \cite{liu2021gpt}, we choose a unidirectional LSTM\footnote{We also tried bidirectional LSTM as the bridge layers in this step. However, with bidirectional LSTM, the continuous prompt model has worse performance. The possible reason for that is that CCTO promt and TTNT promt is independent and these two discrete prompt tokens is also unidirectional in discrete prompt model given the property of auto-regressive GPT-2.} with a ReLU activated two-layer multilayer perceptron (MLP) to convert the classifier outputs to the continuous prompts. As shown in Equation \ref{equ: LSTM bridge layer}, the pooled outputs of CCTO classifier and TTNT classifier are fed into LSTM bridge layers and generate the CCTO and TTNT continuous prompts. The $\emph{cp}^{ccto}$ and $\emph{cp}^{ttnt}$ continuous prompts are used to replace the word embeddings of the CCTO and TTNT prompt tokens in discrete prompt learning respectively, and then append word embeddings of other input tokens to guide the response generation. Because the hidden size of RoBERTa and embedding size of GPT-2 are both $768$, hence the input size and output size of LSTM layer are also set to $768$, and with dropout rate $0.1$.
\begin{equation}
\label{equ: LSTM bridge layer}
    [\emph{cp}^{ccto}, \emph{cp}^{ttnt}] = \text{MLP}(\text{LSTM}([\emph{p}^{ccto}, \emph{p}^{ttnt}]))
\end{equation}

\begin{table*}[h!!!]
\centering
\caption{Automatic metrics of normal chit-chat and task-oriented response generation in the unified GPT-2 model and performance of transition sentence generation in the system-initiated discrete and continuous prompt models. The evaluation on transition sentences is only with human augmented FusedChat dialogues.}
\scalebox{1.0}{
\begin{tabular}{lcccccccc}
    \toprule
    & \multicolumn{2}{c}{Chit-Chat} & \multicolumn{3}{c}{Task-Oriented}  & \multicolumn{2}{c}{Transition Sentence} \\
    \cmidrule(lr{.75em}){2-3}  \cmidrule(lr{.75em}){4-6} \cmidrule(lr{.75em}){7-8}
    
    & \makecell{Distinct-1 \\ (\%)} & \makecell{Distinct-2 \\ (\%)} &
    \makecell{BLEU-4 \\ (\%)} & \makecell{Meteor \\(\%)} & \makecell{BERTScore \\ F1 (\%)} & \makecell{Transition \\ Accuracy (\%)} & \makecell{BLEU-4 \\ (\%)}\\ 
    
   \midrule
   unified GPT-2 & 1.64 &  11.60 & 34.55 &  55.55 &  93.23 & - & - \\
   
   \midrule
   discrete prompt model  &  1.53  &  11.23  &  30.87  &  51.06  &  92.56 & 98.98 & 3.82\\  
   
   \midrule
   continuous prompt model  &  1.55  &  11.20  &  30.25  &  50.27  &  92.51 & 56.95 & 1.54  \\ 

   \bottomrule
\end{tabular}}
\label{tab: the automaic performance of unified generation model, discrete and dcontinuous prompt model.}
\end{table*}

\subsubsection{Continuous Prompt Model}
\label{subsubsec: Initiative Generation Model}

In the continuous prompt model, the fine-tuned RoBERTa classifiers (section \ref{subsubsec: RoBERTa Classifier}) and the discrete prompt GPT-2 (section \ref{subsec: Discrete Prompting}) are connected through the LSTM bridge layers (section \ref{subsubsec: LSTM Bridge Layer}). The architecture of the continuous prompt model is shown in Figure \ref{fig: architecture of unified model, plus discrete prompt learning and continuous prompt learning with pre-trained classifiers.}. The RoBERTa classifiers automatically predict the generation modes, then LSTM layers project the classifier outputs to the continuous prompts and feed them into the word embedding layer of discrete prompt GPT-2. Because RoBERTa classifiers and discrete prompt GPT-2 have already been trained separately, we freeze both RoBERTa classifiers and discrete prompt GPT-2 to only train the LSTM bridge layers in this step.

In order to better train the parameters $\theta^{''}$ of LSTM bridge layers, the first and second prompt predictions are also included besides response generation during training. As shown in Equation \ref{equ: loss funtion in continuous prompt training}, $\textit{ccto}$ prediction could be $\text{[CHIT-CHAT]}$ or $\text{[TASK-ORIENTED]}$ token; $\textit{ttnt}$ prediction could be $\text{[TRANSITION-TURN]}$ or $\text{[NORMAL-TURN]}$ token. In this step, we train the LSTM bridge layers with all turns of human augmented dialogues. The input of RoBERTa is a maximum of $256$ dialogue history tokens, which is aligned with the fine-tuned RoBERTa classifiers. The input of GPT-2 is the past $3$ turns' dialogue context, which is the same as the training of the discrete prompt model. The batch size is $16$, learning rate is $5\mathrm{e}{-5}$ and best model is saved at epoch $10$ with early stopping.
\begin{equation}
\label{equ: loss funtion in continuous prompt training}
    p_{\theta^{''}}(ccto, ttnt, \emph{r})  = 
    \begin{cases}
     \prod_{t=1}^{T} p_{\theta^{''}}(r_{t}|r_{<t}, cp_{ccto}, cp_{ttnt}, \emph{u}, \emph{d}) \\ 
     \quad\quad\quad\quad\quad\quad \text{if task-oriented} \\
     
     \prod_{t=1}^{T} p_{\theta^{''}}(r_{t}|r_{<t}, cp_{ccto}, cp_{ttnt}, \emph{u}) \\
     \quad\quad\quad\quad\quad\quad \text{if chit-chat}
    \end{cases}
\end{equation}

\section{Experimental Results}
\label{sec: Results Comparison and Use Case Study}
This section elaborates on the performance of our models from different perspectives, including automatic metrics comparison and use case studies.

\subsection{Automatic Metrics Comparison}
\label{subsec: Automatic Metrics Comparison}

This section discusses the comparison of automatic metrics between different generation models shown in the Table \ref{tab: the automaic performance of unified generation model, discrete and dcontinuous prompt model.}.

To evaluate generated chit-chat responses, {Distinct-1} and {Distinct-2} \citep{li2016diversity} are used to measure the proportion of distinct unigrams and bigrams in all generated results to indicate diversity. To evaluate generated task-oriented responses, the $\emph{N}$-gram matching metrics, BLEU-4 \citep{papineni2002bleu} and Meteor \citep{banerjee2005meteor} are used to evaluate overall generation quality. In addition, the machine learned automatic metric BERTScore \citep{zhang2019bertscore} is also utilized to evaluate the task-oriented responses. Compared with the unified model, our proposed system-initiated generation models, namely the discrete prompt model and the continuous prompt model, both suffer a small but acceptable loss in automatic metrics for normal chit-chat and task-oriented response generation. This is expected compared to the unified model that can only generate normal chit-chat and task-oriented responses, while a new feature, transitional sentence generation, is activated along with updated parameters in the discrete and continuous prompt models. Therefore, how to extend the initiative capability in a unified conversational model while maintaining its original performance will be our future research direction. 

For transition sentence generation, we propose an automatic metric, \textit{transition accuracy}, to detect whether the responses generated at transition turns include the $\text{[TRANSITION]}$ special token, which represents the proactive transitions in the initiative dialogue models are activated. Adapted from the unified generation model, the discrete prompt model can generate a normal task-oriented or chit-chat response along with a transition sentence at the transition turn with $98.98\%$ transition accuracy. To further verify the effectiveness of prompt tokens in this work, we also trained a discrete prompt model without two prompt tokens. This leads to an extremely low transition accuracy of only $3.05\%$. This is a good evidence that our proposed prompt learning method works and helps the model to be sensitive to dialogue mode transitions. The proposed continuous prompt model has a lower transition accuracy compared to the discrete prompt model even though it can predict generation modes automatically. The possible reason is that the dataset is seriously imbalanced between normal turns and transition turns, i.e., there is only one transition turn in multi-turn FusedChat dialogues. However, the latent high-dimensional continuous prompts show us another transition possibility between task-oriented domains, like a proactive transition from restaurant domain to taxi domain, as the third dialogue example in Table \ref{tab: use cases study to show the ability to generate the transition sentence in different scenarios.}. This also inspires our future work on the system-initiated transitions in a task-oriented multi-domain setting.

\begin{table*}[h!!!]
\centering
\caption{Three different dialogue scenarios with generated transition sentence (highlighted in \textbf{bold}). The [D] means the discrete prompt model, and [C] represents the continuous prompt model.}
\scalebox{0.95}{
\begin{tabular}{cc}
    \toprule
    
    \makecell[c]{dialogue \\ context \\ CC->TO}  & \makecell[l]{\textbf{User:} My friend told me to try some Indian food.
    She said that it would be something that I like, \\ given how I always like my food spicy. \\
    \textbf{System:} Sounds exciting! \\
    \textbf{User:} Yeap. I have never tried it before and I am looking forward to it. }\\
    
    \midrule
    
    \makecell[c]{response at \\ transition turn} & \makecell[l]{\textbf{[D]:} Hope that you have fun then! \textbf{By the way, I can help with the booking of the Indian restaurant.}  \\ 
    \textbf{[C]:} I hope you have fun then. \textbf{I can recommend some Indian restaurants if you need.}} \\
    
    \midrule
    \midrule
    
    \makecell[c]{dialogue \\ context \\ TO->CC} & \makecell[l]{\textbf{User:} I am trying to find a cheap Asian restaurant to have dinner at tonight. What kind of options \\ are available for me? \\
    \textbf{System:} I have two places, how many people and what time? \\
    \textbf{User:} Actually, can you suggest one of them. Can I get their contact info as well?} \\
    
    \midrule
    
    \makecell[c]{response at \\ transition turn} & \makecell[l]{\textbf{[D]:} Sure, I recommend the Dojo Noodle Bar, located at 40210 millers yard city centre, phone 01223363471. \\ \textbf{By the way, are you going with your friends?} \\
    \textbf{[C]:} Dojo Noodle Bar is located at 40210 millers yard city centre, and their phone number is 01223363471. \\ \textbf{Any special reason for having a dinner?}} \\

    \midrule
    \midrule

    \makecell[c]{dialogue \\ context \\ domain->domain}  & \makecell[l]{\textbf{User:} I am looking for suggestions or a cheaper restaurant in the center of town. \\
    \textbf{System:} I have several listings available for you to choose from, is there any \\ preference on the type of food they serve for you? \\
    \textbf{User:} Not really, just get me a table for 3 on Tuesday at 17:00. \\
    \textbf{System:} Would you like me to book you a table at Dojo Noodle bar at 40210 millers \\ yard city centre for Tuesday at 17:00 for 3 people? \\
    \textbf{User:} Yes please. If that time does not work, we can try for 16:00.}\\
    
    \midrule
    
    \makecell[c]{response at \\ transition turn} & \makecell[l]{\textbf{[C]:} I have booked you at Dojo Noodle bar at 16:00 for 3 people. Your reference number is v65s4lw2. \\ \textbf{By the way, do you need help with booking the transportation?}} \\

    \bottomrule

\end{tabular}}

\label{tab: use cases study to show the ability to generate the transition sentence in different scenarios.}
\end{table*}

\subsection{Use Cases Study}
\label{subsec: use case study}

In addition to automatic metric evaluation, we also illustrate several dialogue examples with generated transition sentences by our proposed models in this section.

The first and second dialogue cases in Table \ref{tab: use cases study to show the ability to generate the transition sentence in different scenarios.} show that our proposed discrete prompt and continuous prompt model both can generate a reasonable transition sentence at the transition turn to realize the system-initiated transitions between chit-chat and task-oriented. In dialogue mode transitions from chit-chat to task-oriented, prepended chit-chat generally includes a slot (e.g., ``Indian food'' in the first case in Table \ref{tab: use cases study to show the ability to generate the transition sentence in different scenarios.}), to start with the succeeding task-oriented dialogue and this information can potentially guide transition sentence generation. However, it is more difficult for the system to realize proactive transition from task-oriented to chit-chat, where there is no explicit transition topic to the succeeding chit-chat dialogue. Furthermore, the use cases study also show that the generation of transition sentences to chit-chat interaction is highly dependent on human augmented dataset. The third dialogue case in Table \ref{tab: use cases study to show the ability to generate the transition sentence in different scenarios.} show another transition possibilities between task domains (transition from restaurant to train or taxi domain) in the continuous prompt model. Because most of task-oriented dialogues in MultiWOZ are multi-domain, the continuous prompt model learns the possibility, domain transitions, from latent high-dimensional continuous prompts. This is also an additional advantage of continuous prompt learning compared with discrete prompt learning and points us a future direction for this research.

\section{Conclusion and Outlook}
\label{sec:Conclusion and Future Work}
In this work, we explore the system-initiated transitions between dialogue modes in dialogue systems. Adapted from pre-trained GPT-2, we first build a unified model that can reply to both chit-chat and task-oriented requests. Afterwards, we extend the proactive transitions of the unified model through efficient prompt learning. By prepending only two special discrete tokens, we build a discrete prompt model adapted from unified model. Furthermore, we utilize pre-trained RoBERTa to predict dialogue modes directly from dialogue history and generate continuous prompts through LSTM bridge layers for the continuous prompt model. Both initiative models show promising performance in proactive switching between chit-chat and task-oriented dialogues. 

We did not perform inter-rater reliability scores on the human augmentation task (section \ref{sec: dataset collection}), nor human evaluation (section \ref{sec: Results Comparison and Use Case Study}) for performance comparison. These limitations will be addressed in future work.

The study of proactive transitions in a unified conversational model is a new problem. The promising performance of this work motivates us to continue this research.
The proposed discrete prompt model can better control the transitions initiated by the system, while the continuous prompt model can learn more potential system-initiated possibilities from a latent high-dimensional space.
However, both suffer from the problem of insufficient training data, especially for the continuous prompt model.
Based on current experiment results, we are also interested in exploring the proactive transitions from chit-chat to task-oriented and from task-oriented to chit-chat in a unified dialogue model separately, because the transition sentence generation in these two scenarios are totally different.
In FusedChat, only one turn in each multi-turn dialogue is labelled as the transition turn. In real dialogue scenarios, it might be much more complicated. Therefore, multiple rounds of proactive transitions in one dialogue is a more challenging research topic.




\bibliographystyle{ACM-Reference-Format}
\bibliography{sample-base}

\appendix





\end{document}